\documentclass{article}

\usepackage[preprint]{neurips_2024}

\usepackage[utf8]{inputenc} 
\usepackage[T1]{fontenc}    
\usepackage{hyperref}       
\usepackage{url}            
\usepackage{booktabs}       
\usepackage{amsfonts}       
\usepackage{nicefrac}       
\usepackage{microtype}      
\usepackage{xcolor}         
\usepackage{graphicx}
\usepackage{natbib}
\usepackage{dsfont}
\setcitestyle{numbers,square,comma}
\usepackage{multirow}
\usepackage{wrapfig}
\usepackage{pgfplots}
\pgfplotsset{compat=1.18}
\usepgfplotslibrary{groupplots}

\usepackage{amsmath}
\usepackage{amsthm}
\usepackage{colortbl}
\usepackage{bbding}
\usepackage{subcaption}
\usepackage{multicol}
\usepackage{sidecap}
\usepackage[linesnumbered,ruled,vlined]{algorithm2e}

\title{Semantic-Rearrangement-Based Multi-Level Alignment for Domain Generalized Segmentation}

\author{%
\textbf{
Guanlong Jiao$^{1}$,
Chenyangguang Zhang$^{1}$, 
Haonan Yin${^1}$,
Yu Mo${^1}$,
Biqing Huang${^1}{^\dag}$
},\\
\textbf{
Hui Pan$^{2}$,
Yi Luo$^{2}$,
Jingxian Liu$^{2}$
}\\
\textsuperscript{1}Tsinghua University,
\textsuperscript{2}Lenovo Research
\\
\tt\small{\{jgl22@mails., hbq@\}tsinghua.edu.cn}
\thanks{\dag Corresponding author.}
}

\begin{document}

\maketitle

\vspace{-0.04cm}

\begin{abstract}

Domain generalized semantic segmentation is an essential computer vision task, for which models only leverage source data to learn the capability of generalized semantic segmentation towards the unseen target domains.
Previous works typically address this challenge by global style randomization or feature regularization.
In this paper, we argue that given the observation that different local semantic regions perform different visual characteristics from the source domain to the target domain, methods focusing on global operations are hard to capture such regional discrepancies, thus failing to construct domain-invariant representations with the consistency from local to global level.
Therefore, we propose the \textbf{S}emantic-\textbf{R}earrangement-based \textbf{M}ulti-Level \textbf{A}lignment (SRMA) to overcome this problem.
SRMA first incorporates a Semantic Rearrangement Module (SRM), which conducts semantic region randomization to enhance the diversity of the source domain sufficiently.
A Multi-Level Alignment module (MLA) is subsequently proposed with the help of such diversity to establish the global-regional-local consistent domain-invariant representations.
By aligning features across randomized samples with domain-neutral knowledge at multiple levels, SRMA provides a more robust way to handle the source-target domain gap.
Extensive experiments demonstrate the superiority of SRMA over the current state-of-the-art works on various benchmarks. 

\end{abstract}

\begin{figure}
    \centering
    \includegraphics[width=0.87\linewidth]{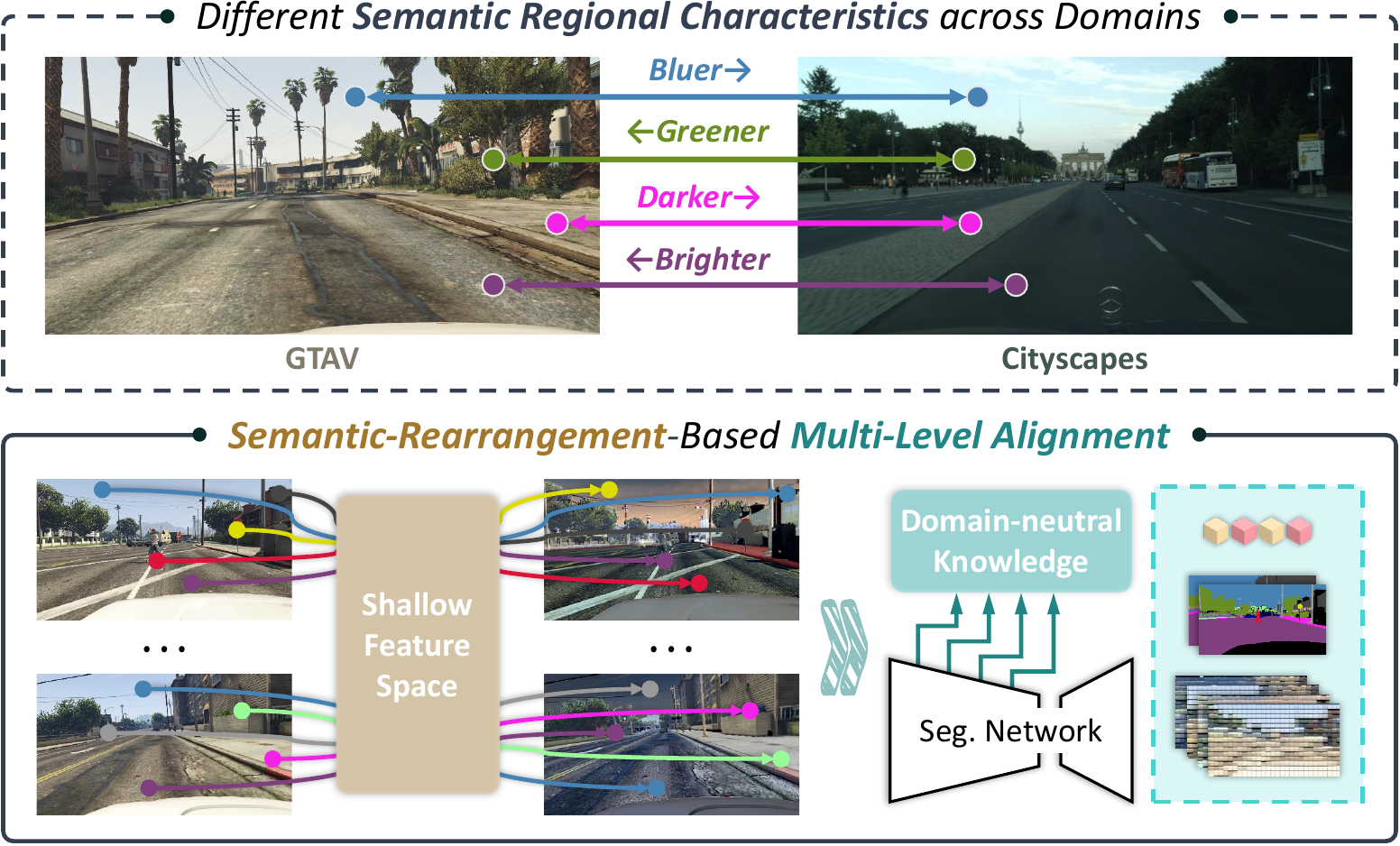}
    \vspace{-0.1cm}
    \caption{Illustration of different semantic regional characteristics across domains (top) and the proposed semantic-rearrangement-based multi-level alignment (bottom). 
    There are visual style discrepancies including color and brightness among the individual semantic regions in different domains. 
    To accommodate this domain variance, we propose the semantic rearrangement module to enrich the semantic regional characteristics in the source domain and exploit domain-invariant representations from the randomized semantic regional characteristics via multi-level alignment.
    }
    \vspace{-0.4cm}
    \label{fig:teaser}
\end{figure}

\section{Introduction}

Deep-neural-network-based models have yielded promising results in many computer vision tasks \cite{krizhevsky2012imagenet,simonyan2014very,badrinarayanan2017segnet,yin2023lafite,zhang2023moho}, especially in semantic segmentation \cite{chen2014semantic,girshick2015fast,he2017mask,chen2017deeplab,cheng2022masked,Kirillov_2023_ICCV}.
These segmentation methods benefit from large-scale annotated training data.
However, their performance degrades dramatically when evaluated on unseen out-of-distribution samples due to the severe source-target domain gap.
To tackle this problem, a more practical but challenging setting of Domain Generalization (DG) has been introduced in semantic segmentation \cite{pan2018two,yue2019domain,choi2021robustnet}, which only leverages source samples and aims at learning a generalized model to cope with different unseen domains. 

In Domain Generalized Semantic Segmentation (DGSS), the inaccessible nature of the target domain means the source-target domain gap is unidentified. 
Hence, the principal concept of DGSS aims to diversify the training data or eliminate domain-specific characteristics to facilitate the learning of domain-invariant models. 
To increase the diversity of training data, some approaches augment the source domain by performing a photometric transformation on the entire image \cite{wu2022siamdoge,xu2022dirl,chattopadhyay2023pasta,kim2023texture} or a global style transfer in the feature space \cite{yue2019domain,lee2022wildnet,zhao2022style}.
Furthermore, to learn domain-invariant models, most previous works incorporate feature regularization techniques \cite{pan2018two,choi2021robustnet,xu2022dirl} that focus on reducing the domain-specific information contained in the global statistics, such as the covariance and the mean of overall pixels.

Generally speaking, the majority of the previous works in the two streams introduced above devise various methods to increase global diversity or constrain features to possess certain global attributes.
However, such analyses take into account only a narrow fraction of the domain characteristics, neglecting diverse cross-domain expressions in individual semantic regions.
For instance, as shown in Fig.\ref{fig:teaser}, some semantic regions exhibit vastly different characteristics across domains.
GTAV \cite{richter2016playing} consistently presents greener \textit{vegetation} and brighter \textit{road}, while Cityscapes \cite{cordts2016cityscapes} presents bluer \textit{sky} and darker \textit{sidewalk}.
Consequently, different semantic regions perform varying visual characteristics when transferred from the source domain to the target domain.

Previous methods typically neglect such phenomenon in two aspects.
\textbf{First}, since their domain randomization is conducted via the global style transfer, the different regional expressions in each semantic region are hard to capture.
This insufficient randomization limits the model's generalization capability when facing complex scenarios where the regional styles of the target domain are hard to simulate through simple global augmentations.
To this end, we introduce a semantic-rearrangement-based randomization approach which increases the diversity of semantic regional expressions.
\textbf{Second}, existing global feature regularization techniques do not align such regional discrepancies between the source domain and the target domain, resulting in deficiencies in obtaining the generalization capability for individual semantic regions across domains.
Therefore, we propose the multi-level alignment to establish the global-regional-local consistent domain-invariant representations for DGSS.

Motivated by the aforementioned analysis, we introduce a novel Semantic-Rearrangement-based Multi-Level Alignment (SRMA) framework to learn the global-regional-local consistent domain-invariant representations with the help of semantic-regional-rearranged samples.
First, we introduce a \textbf{Semantic Rearrangement Module} (SRM) that conducts semantic region randomization to enhance the diversity of the source domain.
Specifically, we regard the channel-wise mean and standard deviation as style features of each semantic region and utilize these style features to generate diverse samples based on their randomly weighted combinations, thus providing varied semantic regional characteristics.
Equipped with such regional randomized features, we subsequently propose a \textbf{Multi-Level Alignment module} (MLA) to construct the global-regional-local consistent domain-invariant representations.
MLA exploits global, regional, and local level alignments to help models understand domain-invariant attributes by aligning the deep features at multiple levels.
A fixed pre-trained feature extractor is adopted in MLA to provide domain-neutral knowledge for alignments.
We impose the features extracted by our segmentation model to be consistent with the pre-trained domain-neutral features at multiple levels to avoid overfitting the source domain.
Specifically, we first align global centers to learn unbiased representations at the global level.
Then, to fully leverage the regional diversity provided by SRM, we constrain the semantic regional centers to be consistent towards the domain-neutral features.
Additionally, MLA further includes a local level alignment to provide the local domain-invariant expressions following \cite{zhao2022style,hoyer2022daformer}.
Such multiple constraints provide fine-grained guidance into domain invariance, contributing to the robustness of DGSS models.

Our main contributions are as follows.
\textbf{First}, based on the observation that each individual semantic region has different characteristics across domains, we propose the semantic rearrangement module for semantic region randomization to enhance the regional diversity of the source domain.
\textbf{Second}, we propose the multi-level alignment to construct the global-regional-local consistent domain-invariant representations for DGSS. 
\textbf{Third}, we conduct domain invariance quantitative and qualitative analyses with extensive experiments on multiple DGSS benchmarks, demonstrating the effectiveness of SRMA which outperforms state-of-the-art methods on various settings.

\section{Related Works}

\textbf{Domain Adaption and Generalization.} 
Domain adaption (DA) and domain generalization (DG) seek to narrow the source-target domain gap.
DA methods learn task-specific knowledge through supervised training on the source domain and adapt to target characteristics via unsupervised learning on the target domain \cite{long2016unsupervised,tzeng2017adversarial,hoyer2022hrda,xie2023sepico}.
In contrast to DA, DG involves moving away from reliance on unlabeled target domain data \cite{gan2016learning}.
DG mitigates the performance degradation of evaluating data from unseen domains by learning domain-invariant representations on the source domain \cite{muandet2013domain,li2018learning,seo2020learning,nam2021reducing}.
To better realize this, various DG approaches perform data augmentation \cite{zhou2020learning,nam2021reducing}, metric-learning \cite{motiian2017unified,wang2020learning}, and feature regularization \cite{pan2018two,seo2020learning} on image classification tasks.
However, limited by the fact of having only image-level annotations, most of the methods attempt domain generalization only from the global characteristics of images.

\noindent \textbf{Domain Generalized Semantic Segmentation.}
Domain generalized semantic segmentation (DGSS) is a relatively innovative task that maintains the same challenging setting as DG and demands additional attention to fine-grained semantic information.
Recent approaches can be divided into two categories: instance-level and feature-level approaches.
Instance-level methods incline to generate new training samples using \textit{image augmentation} \cite{yue2019domain,wu2022siamdoge,kim2023texture,chattopadhyay2023pasta} or \textit{feature randomization} \cite{zhao2022style,lee2022wildnet}.
However, most of these methods perform globally consistent randomization, taking only global attributes of the image into account.
We point out that individual semantic regions tend to have different characteristics across domains and propose the semantic rearrangement module (SRM) to diversify such characteristics.

Feature-level methods attempt to perform \textit{feature regularization} to learn domain-invariant representations.
Some approaches impose constraints on the feature statistics in order to obtain more generalized representations \cite{pan2018two,choi2021robustnet}, but they only consider the global information that is insufficient for DGSS.
Some methods learn the local consistency \cite{zhao2022style,hoyer2022daformer}, but fail to avoid models' lacking understanding of domain-invariant attributes.
SAN-SAW \cite{peng2022semantic} seeks to design semantic-aware feature regularization modules, but does not avoid their overfitting to the source domain.
Some other methods try to transfer the out-of-distribution samples back to the source distribution \cite{kim2022pin,huang2023style}, but due to the invisibility and complexity of the target domains, it may be quite hard to match them well with the source distribution.
To tackle these drawbacks, we propose the multi-level alignment (MLA) for learning global-regional-local consistent domain-invariant representations from multiple levels.
Meanwhile, inspired by the methods related to knowledge distillation \cite{zhao2022style,hoyer2022daformer}, we utilize the domain-neutral knowledge contained in the pre-trained model as guidance so as to prevent models from overfitting.

\begin{figure*}
    \centering
    \includegraphics[width=0.99\linewidth]{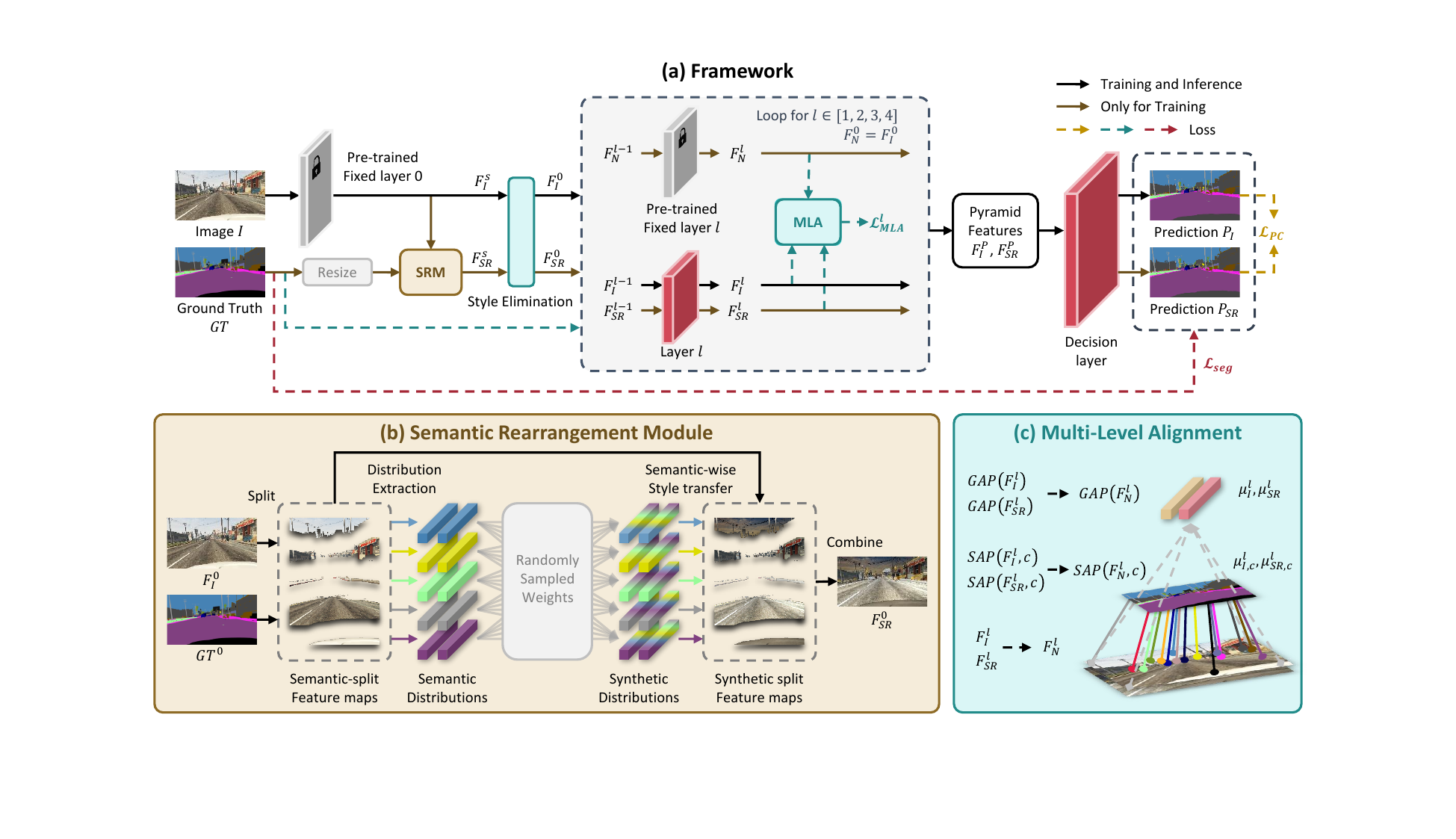}
    \vspace{-0.1cm}
    \caption{Overview of our proposed SRMA.
    (a) Overall framework. (b) Semantic Rearrangement Module (SRM). (c) Multi-Level Alignment (MLA).
    \textit{GAP} and \textit{SAP} indicate global average pooling and semantic average pooling, respectively. 
    We represent shallow features visually using the images and the randomized images shown above are reconstructed from the rearranged shallow features via a decoder used only for visualization.
    }
    \vspace{-0.1cm}
    \label{fig:framework}
\end{figure*}

\section{Method}

\subsection{Overview}
The overall framework of our proposed method is depicted in Fig.\ref{fig:framework}a, consisting of two main components, \textit{i.e.}, a Semantic Rearrangement Module (SRM) and a Multi-Level Alignment constraint (MLA).
Among them, SRM utilizes only input samples themselves for randomization, while MLA contains a parameter-free style elimination and three alignment constraints ($\mathcal{L}_{MLA}$).
We leverage SRM to yield randomized samples in feature space and employ MLA to further learn domain-invariant representations from global, regional, and local levels at deep layers. 
These two components are mainly involved only in the training stage, with just the style elimination playing a role in both the training and inference stages.
In the end, a prediction consistency constraint ($\mathcal{L}_{PC}$) is employed to help learn the task-specific domain insensitivity.

Preliminarily, we briefly introduce the network architecture to make the following descriptions clearer.
We leverage the hierarchical neural network (\textit{e.g.} ResNet \cite{he2016deep}, ShuffleNet \cite{ma2018shufflenet}, MobileNet \cite{sandler2018mobilenetv2}) as the feature extractor and divide it into five layers.
Taking ResNet as an example, we regard the first Conv-BN-ReLU-MaxPool layer as \textit{layer} $\textit{0}$, and the subsequent stage $1\sim4$ as \textit{layer} $\textit{1}\sim\textit{4}$.
In the training stage, SRM provides a semantic-regional-rearranged feature map for each input sample after layer $0$.
Afterward, MLA normalizes both input and randomized samples as well as helps layer $1\sim4$ to learn domain-invariant representations effectively. 
Furthermore, we fix layer $0$ with the pre-trained parameters to avoid its overfitting to the source domain.
Hereafter, we refer to the features extracted by layer $0$ as \textit{shallow features}.

\subsection{Semantic Rearrangement Module (SRM)}

\begin{wrapfigure}{r}{7.0cm}
    \centering
    \vspace{-0.5cm}
    \includegraphics[width=0.99\linewidth]{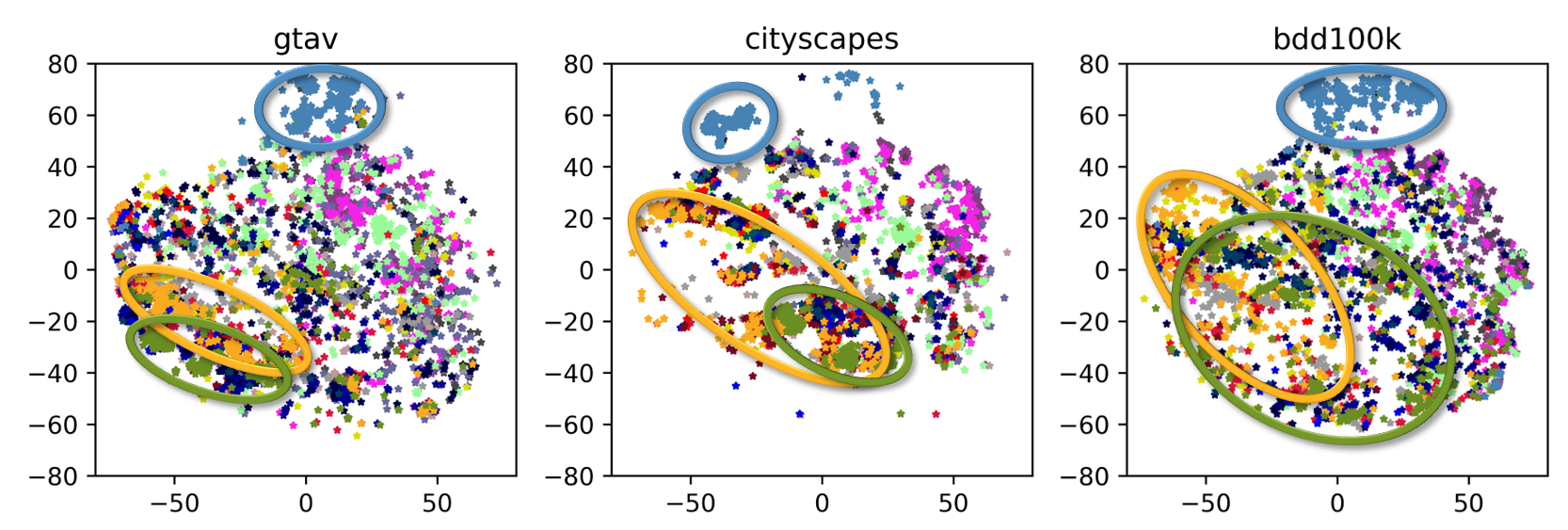}
    \vspace{-0.5cm}
    \caption{
    Visualization of shallow features in different semantic regions across domains. 
    These features are extracted by layer $0$ of ImageNet pre-trained ResNet-50 and visualized by T-SNE dimensionality reduction.
    We circle the regions where \textit{sky} (\textcolor[RGB]{70,130,180}{blue}), \textit{traffic light} (\textcolor[RGB]{250,170,30}{yellow}), and \textit{vegetation} (\textcolor[RGB]{107,142,35}{green}) are concentrated, respectively.
    }
    \vspace{-0.3cm}
    \label{fig:layer0_vis}
\end{wrapfigure}

Our SRM aims to mitigate the impact of different semantic regional characteristics between the source domain and the target domain on semantic segmentation by exposing the model to a wider variety of samples in the training stage.
Preliminarily, as shown in Fig.\ref{fig:teaser}, it is observed that different semantic regions have their specificities across domains. 
Such domain characteristics are not captured at once by the global statistics alone.
To ascertain how these semantic regional characteristics are reflected in the neural network, we visualize the T-SNE \cite{van2008visualizing} embedding of shallow features which mainly contain the style information \cite{pan2018two}, as shown in Fig.\ref{fig:layer0_vis}.
We use different colored points to indicate the feature means of different categories in each sample and compare the distributions of individual semantics across domains.
Take the circled categories as examples, 
we find that the \textit{sky}-concentrated region in Cityscapes \cite{cordts2016cityscapes} barely intersects with its GTAV \cite{richter2016playing} and BDD100K \cite{yu2020bdd100k} counterparts.
Furthermore, there are significant differences between the locations of the regions where \textit{traffic light} / \textit{vegetation} are concentrated in GTAV and the corresponding regions in Cityscapes, whereas the distributions of these two categories are more decentralized in BDD100K.
This implies that the regional characteristics of individual semantics differ across domains, which can make features of different categories indistinguishable and cannot be mitigated by global randomization.
Moreover, such differences can be reflected in the semantic regional styles of shallow features, hence SRM conducts randomization at a shallow layer.

Since shallow layers of the neural network provide more style information than content information \cite{pan2018two}, we leverage SRM right after layer $0$ to avoid corrupting the semantic attributes.
Given the shallow features provided by layer $0$ and the corresponding resized ground truth maps, SRM rearranges the semantic regional styles and provides randomized synthetic shallow features as shown in Fig.\ref{fig:framework}b.
We denote the input image as $I \in \mathcal{R}^{3 \times H \times W}$ and the ground truth of semantic segmentation as $GT \in \mathcal{N}^{H \times W}$, where $H$ and $W$ indicate the length and width of the image, respectively.
The feature extracted by layer $l \in \{1, 2, 3, 4\}$ is denoted as $F^l_I \in \mathcal{R}^{D_l \times H_l \times W_l}$ with $D_l$ indicates the channel dimension.
The ground truth resized to the size of feature map $F^l_I$ by the nearest neighbor interpolation is denoted as $GT^l \in \mathcal{N}^{H_l \times W_l}$.
Specifically, we denote the shallow feature as $F^s_I$ and its corresponding resized ground truth as $GT^s$.
SRM first obtains semantic-split feature maps $F^s_{I,c}$ using the shallow feature map $F^s_I$ and the resized ground truth $GT^s$,
\begin{equation}
    F^s_{I,c} = \{ F^s_{I, (h, w)} ~|~ GT^s_{(h, w)} = c \}^{H_0, W_0}_{h, w=1},
\end{equation}
where $F^s_{I, (h, w)}$ denotes the feature vector at $(h, w)$ on the feature map $F^s_I$ and $c$ denotes the serial number of the semantic category.
Then SRM extracts semantic distributions $(\mu^s_{I,c}, \sigma^s_{I,c}) = (\mu(F^s_{I,c}), \sigma(F^s_{I,c}))$, where $\mu(*)$ and $\sigma(*)$ denotes the channel-wise mean and standard deviation.
Subsequently, we sample a set of combination weights $W_c=[\omega^c_1, ..., \omega^c_C]$ from  Dirichlet distribution $Dir([\alpha_1, ..., \alpha_C])$, where $C$ denotes the total number of semantic categories.
SRM performs a weighted summation of semantic distributions to generate synthetic distributions $(\mu^s_{SR,c}, \sigma^s_{SR,c})$ as follows:
\begin{equation}
\begin{aligned}
    \mu^s_{SR,c} = \sum_{i=1}^{C} \omega^c_i \mu^s_{I,i} ,~~ \sigma^s_{SR,c} = \sum_{i=1}^{C} \omega^c_i \sigma^s_{I,i}.
\end{aligned}
\end{equation}
To transfer the synthetic distributions to the feature map while retaining the content information, we adopt AdaIN \cite{huang2017arbitrary} to gain synthetic semantic-split feature maps $F^s_{SR,c}$, 
\begin{equation}
    F^s_{SR,c} = \sigma^s_{SR,c} \frac{(F^s_{I,c} - \mu^s_{I,c})}{\sigma^s_{I,c}} + \mu^s_{SR,c},
\end{equation}
and combine $[F^s_{SR,1}, ..., F^s_{SR,C}]$ into the semantic rearrangement feature map $F^s_{SR}$ afterwards.
Thus, these semantic rearrangement feature maps serve as additional semantic regional characteristics, facilitating the model to mitigate the impact of the source-target domain gap.

\subsection{Multi-Level Alignment (MLA)}
\label{sec:mla}

To exploit domain-invariant features, previous approaches incline to perform feature regularization techniques \cite{yue2019domain,choi2021robustnet,zhao2022style}. 
However, most of these methods only focus on feature attributes at a particular level, either global or local, which results in them acquiring domain-invariant features with only limited contribution to DGSS.
To this end, we propose MLA, which constrains deep features at the global, regional, and local levels to further exploit domain-invariant representations with the help of rearranged features provided by SRM.

Since SRM only performs semantic region randomization, we consider it necessary to further mitigate the global bias of representations across domains to avoid neglecting the understanding of global domain invariance.
We first introduce a style elimination in MLA to eliminate the global style of shallow features.
By leveraging a parameter-free instance normalization \cite{pan2018two} as the style elimination, MLA obtains the global-style-insignificant features ${F^0_I}$ and ${F^0_{SR}}$,
\begin{equation}
    {F^0_k} = \frac{{F^s_k} - \mu({F^s_k})}{\sigma({F^s_k})}, ~~ k \in \{I, SR\}.
\end{equation}

Subsequently, given the features $F^{l}_I$ and $F^{l}_{SR}$ extracted by layer $l$ from these global-style-insignificant features, MLA yields their global centers through global average pooling (GAP).
Then MLA performs the global level alignment by minimizing the distance between the centers of segmentation features and those of domain-neutral features $F^l_N$ extracted by the fixed pre-trained layer $l$:
\begin{equation}
    \mathcal{A}^l_{g} = \sum_{k \in \{I, SR\}} \frac{\Vert \mu^l_k - GAP(F^l_N) \Vert_2^2}{D_l}, ~~
    \mu^l_k = GAP(F^l_k) = \mu(F^l_k), ~~ 
    k\in\{I, SR\}.
\end{equation}

Subsequently, to capitalize on the semantic regional diversity provided by SRM, MLA performs alignment at the regional level.
MLA fetches the semantic regional centers of features through semantic average pooling (SAP) and takes the weighted sum of the distances from each semantic center to the corresponding domain-neutral one as the regional level alignment constraint:
\begin{equation}
    \mathcal{A}^l_{r} = \sum_{k \in \{I, SR\}} \sum_{c=1}^C \omega_c \frac{\Vert \mu^l_{k, c} - SAP(F^l_N, c) \Vert_2^2}{D_l}, ~~
    \mu^l_{k, c} = SAP(F^l_k, c) = \mu(F^l_{k, c}), ~~ 
    k \in \{I, SR\}.
\end{equation}
where $F^l_{k, c} = \{ F^l_{k, (h, w)} ~|~ GT^l_{(h, w)} = c \}_{h, w=1}^{H_l, W_l}$ and $\omega_c = \frac{ \lvert F^l_{I,c} \rvert }{H_l W_l}$ denotes the ratio of pixels with category $c$ in $GT^l$.

Furthermore, some recent works adopt learning local consistency as a regularization technique \cite{zhao2022style,hoyer2022daformer}, which exploits patch-level domain-invariant information but tends to neglect regional domain-invariant information when used in isolation.
We consider that local expressions are essential to DGSS, thus incorporating the local level alignment into the proposed MLA:
\begin{equation}
\begin{aligned}
    \mathcal{A}^l_{l} = \sum_{k \in \{I, SR\}} \frac{1}{H_l W_l} \sum_{h,w=1}^{H_l, W_l} \frac{ \Vert F^l_{k, (h,w)} - F^l_{N, (h,w)} \Vert_2^2 }{D_l}, 
\end{aligned}
\end{equation}

Collectively, as shown in Fig.\ref{fig:framework} (c), MLA contains global, regional, and local level alignment constraints, focusing on cross-domain feature unbiasedness at the global level, alignment at the regional level, and feature consistency at the local level.
It can be represented as a weighted sum of these three alignments:
\begin{equation}
    \mathcal{L}^l_{MLA} = \mathcal{A}^l_{g} + \mathcal{A}^l_{r} + \mathcal{A}^l_{l},
\end{equation}

Moreover, since each deep layer of the neural network has different degrees of global style / content information and different local receptive fields, we attempt to constrain all the deep layers with a set of weights $\lambda_{MLA}$:
\begin{equation}
    \lambda_{MLA}\mathcal{L}_{MLA} = \sum_{l \in \{1, 2, 3, 4\}} \lambda_{MLA}^l \mathcal{L}^l_{MLA}.
\end{equation}

\subsection{Training Objective}

\noindent\textbf{Prediction Consistency Loss.}
Previously, MLA focuses on learning domain-invariant characteristics in features.
Prediction consistency further provides regularization of task-specific knowledge to improve the robustness of the segmentation model to the source-target domain gap.
We leverage the Jensen-Shannon Divergence between the posterior probabilities $P_I$ and $P_{SR}$ as a consistency metric inspired by the instance-level based methods \cite{yue2019domain,peng2021global,zhao2022style}.
The prediction consistency loss can be obtained by:
\begin{equation}
    \mathcal{L}_{PC} = \frac{1}{2} ( KL(P_I \Vert P_{JS}) + KL(P_{SR} \Vert P_{JS}) ) ,
\end{equation}
where $KL$ denotes the KL Divergence and $P_{JS} = (P_I + P_{SR}) / 2$.

\noindent\textbf{Overall Training Objective.}
We use $\mathcal{L}_{task}$ to denote the segmentation task loss (\textit{e.g.}, a per-pixel cross-entropy loss), and subsequently indicate the segmentation loss of the image and the semantic-regional-rearranged sample by $\mathcal{L}_{task, I}$ and $\mathcal{L}_{task, SR}$, respectively.
The overall training objective is described as:
\begin{equation}
    \mathcal{L} = \frac{1}{2} (\mathcal{L}_{task, I} + \mathcal{L}_{task, SR}) + \lambda_{MLA}\mathcal{L}_{MLA} + \lambda_{PC} \mathcal{L}_{PC},
\end{equation}
where $\lambda_{PC}$ is set to 10 and $\lambda_{MLA}$ is set to a linearly increasing set $[0.4, 0.6, 0.8, 1]$ with respect to the layer depth empirically.
The concentration parameters of Dirichlet distribution are all set to $2^{-6}$.

\section{Experiments}

\subsection{Experimental Setup}

\noindent\textbf{Datasets.}
We conduct the experiments on five different semantic segmentation datasets including synthetic datasets and real-world datasets to evaluate the generalization capability of our method.
In terms of synthetic datasets, GTAV (G) \cite{richter2016playing} consists of 24966 images with a resolution of 1914$\times$1052 captured from the GTA-V game engine.
Synthia (S) \cite{ros2016synthia} contains 9400 images of virtual urban scenes with a resolution of 1280$\times$760.
As real-world datasets, Cityscapes (C) \cite{cordts2016cityscapes} consists of 5000 images with a resolution of 2048$\times$1024 from 50 different cities mainly in Germany.
BDD100K (B) \cite{yu2020bdd100k} contains 10000 images from the US with a resolution of 1280$\times$720.
Mapillary (M) \cite{neuhold2017mapillary} contains 25000 images with an average resolution of 1914$\times$1080 collected from all around the world.

\noindent\textbf{Network architecture.}
Following previous works \cite{choi2021robustnet,zhao2022style}, we adopt DeepLabV3+ \cite{chen2018encoder} as the segmentation architecture.
We conduct experiments using ResNet-50 \cite{he2016deep}, ShuffleNetV2 \cite{ma2018shufflenet} and MobileNetV2 \cite{sandler2018mobilenetv2} as backbones, and all of them in our experiments are initialized with ImageNet \cite{deng2009imagenet} pre-trained weights.
We fix layer $0$ and apply the parameter-free style elimination after layer $0$.

\noindent\textbf{Implementation Details.}
We adopt a SGD optimizer with the momentum 0.9 and weight decay $5\times10^{-4}$.
An initial learning rate is set to 0.005 for the encoder and 0.01 for the decoder.
We utilize the polynomial learning rate scheduling \cite{liu2015parsenet} with the power of 0.9 and train the models on NVIDIA RTX 3090 GPU.
Color jittering, Gaussian blur, random flipping, and random cropping of 768$\times$768 are adopted for training following \cite{choi2021robustnet}.
A total of 19 semantic categories are used for training and validation.
We utilize the mean Intersection over Union (mIoU) averaged over all categories as the evaluation metric.

\newcommand{\secondmark}[1]{\underline{#1}}
\newcommand{\scaletable}[1]{\scalebox{0.85}{#1}}

\subsection{Comparison with DGSS methods}

We extensively compare our SRMA against state-of-the-art DGSS methods.
These methods can be categorized into various groups, including 
\textit{feature regularization} (IBN-Net \cite{pan2018two}, ISW \cite{choi2021robustnet}, SAN-SAW \cite{peng2022semantic}, PintheMem \cite{kim2022pin}, DIRL \cite{xu2022dirl}, SPC-Net \cite{huang2023style}), 
\textit{image augmentation} (DRPC \cite{yue2019domain}, PASTA \cite{chattopadhyay2023pasta}, TLDR \cite{kim2023texture}) and 
\textit{feature randomization} (WildNet \cite{lee2022wildnet}, SHADE \cite{zhao2022style}).
Among these methods, DRPC, WildNet, and TLDR utilize the extra ImageNet \cite{deng2009imagenet} dataset as an auxiliary, we mark them with $^{\S}$.

\begin{table}
    \centering
    \scaletable{
    \begin{tabular}{l|ccc|c||ccc|c}
        \toprule
        \multirow{2}{*}{Methods}    & \multicolumn{4}{c||}{ResNet-50} & \multicolumn{4}{c}{ResNet-101} \\
            & C & B & M & Avg. & C & B & M & Avg. \\
        \midrule
            Baseline  & 28.95 & 25.14 & 28.18 & 27.42   & 32.97 & 30.77 & 30.68 & 31.47 \\
            IBN-Net \cite{pan2018two} (ECCV'18)               & 33.85 & 32.30 & 37.75 & 34.63   & 37.37 & 34.21 & 36.81 & 36.13 \\
            DRPC$^{\S}$ \cite{yue2019domain} (ICCV'19)        & 37.42 & 32.14 & 34.12 & 34.56    & 42.53 & 38.72 & 38.05 & 39.77 \\  
            ISW \cite{choi2021robustnet} (CVPR'21)            & 36.58 & 35.20 & 40.33 & 37.37   & 37.20 & 33.36 & 35.57 & 35.38 \\
            WildNet$^{\S}$ \cite{lee2022wildnet} (CVPR'22)    & 44.62 & 38.42 & 46.09 & 43.04    & 45.79 & 41.73 & 47.08 & 44.87 \\
            PintheMem \cite{kim2022pin} (CVPR'22)             & 41.00 & 34.60 & 37.40 & 37.67   & -     & -     & -     & - \\
            SAN-SAW \cite{peng2022semantic} (CVPR'22)         & 39.75 & 37.34 & 41.86 & 39.65   & 45.33 & 41.18 & 40.77 & 42.43 \\
            DIRL \cite{xu2022dirl} (AAAI'22)                  & 41.04 & 39.15 & 41.60 & 40.60   & -     & -     & -     & - \\
            SHADE \cite{zhao2022style} (ECCV'22)              & 44.65 & 39.28 & 43.34 & 42.42   & 46.66 & 43.66 & 45.50 & 45.27 \\
            SPC-Net \cite{huang2023style} (CVPR'23)           & 44.10 & 40.46 & 45.51 & 43.36   & -     & -     & -     & -  \\
            PASTA \cite{chattopadhyay2023pasta} (ICCV'23)     & 44.12 & 40.19 & \secondmark{47.11} & 43.81   & 45.33 & 42.32 & 48.60 & 45.42 \\
            TLDR$^{\S}$ \cite{kim2023texture} (ICCV'23)       & \secondmark{46.51} & \textbf{42.58} & 46.18  & \secondmark{45.09}   & \secondmark{47.58} & \secondmark{44.88} & \secondmark{48.80} & \secondmark{47.09} \\
        \rowcolor{gray!20} \textbf{Ours}            & \textbf{47.05} & \secondmark{41.55} & \textbf{47.96} & \textbf{45.52}     & \textbf{48.38} & \textbf{45.02} & \textbf{49.56} & \textbf{47.65} \\
        \bottomrule
    \end{tabular}}
    \caption{Mean IoU (\%, higher is better) comparison between existing DGSS methods trained on GTAV and evaluated on Cityscapes (C), BDD100K (B), and Mapillary (M) with ResNet-based backbones. 
    The best and second best results are \textbf{highlighted} and \secondmark{underlined}, respectively. 
    $^{\S}$ denotes using extra data during training.
    The results of previous methods come from their public reports.
    }
    \label{tab:gtav}
    \vspace{-0.4cm}
\end{table}

Tab.\ref{tab:gtav} shows the generalization performance of models trained on GTAV using ResNet-based backbones.
We evaluate models on three real-world datasets, including Cityscapes, BDD100K, and Mapillary, and also report the average value of mIoU on these datasets.
SRMA achieves an average mIoU of 45.52\% on the three real-world datasets when using ResNet-50, which not only beats models using extra training data (\textit{e.g.}, DPRC, WildNet) or parameters (\textit{e.g.}, DIRL, SPC-Net), but also outperforms the previous best method TLDR by 0.43\% mIoU on average.
Additionally, in terms of both individual unseen target domains and average performance, our method shows consistently superior performances using ResNet-101, demonstrating the superiority of our method while supporting its use in even more complex scenarios.
Furthermore, as shown in Tab.\ref{tab:cityscapes}, SRMA also achieves an average mIoU of 43.67\% using ResNet-50 on the domain generalization setting from a small-scale real-world dataset to other domains (\textit{i.e.}, trained on Cityscapes and validated on BDD100K, Synthia, and GTAV), improving over the state-of-the-art method by 0.82\% mIoU on average. 
As a semantic segmentation training set, Cityscapes is small and lacks diversity.
Our performance on this setting indicates the effectiveness of SRMA when facing the less varied source domain.
We further compare the generalization performance of models using lightweight backbones (\textit{i.e.}, ShuffleNet \cite{ma2018shufflenet} and MobileNet \cite{sandler2018mobilenetv2}) and other architecture (\textit{i.e.}, Mask2Former \cite{cheng2022masked}) in Appendix \ref{sec:lightweight} and \ref{sec:m2f}, respectively, which consistently demonstrating the superiority of SRMA.

\begin{table}
    \centering
    \scaletable{
    \begin{tabular}{l|ccc|c}
        \toprule
        ~~~Methods~~~ & ~~~~~B~~~~~ & ~~~~~S~~~~~ & ~~~~~G~~~~~ & ~~~Avg.~~~ \\
        \midrule
        Baseline & 44.96 & 23.29 & 42.55 & 36.93 \\
        IBN-Net \cite{pan2018two} (ECCV'18)               & 48.56 & 26.14 & 45.06 & 39.92 \\
        DRPC$^{\S}$ \cite{yue2019domain} (ICCV'19)        & 49.86 & 26.58 & 45.62 & 40.69 \\ 
        ISW \cite{choi2021robustnet} (CVPR'21)            & 50.73 & 26.20 & 45.00 & 40.64 \\
        WildNet$^{\S}$ \cite{lee2022wildnet} (CVPR'22)    & 50.94 & \secondmark{27.95} & 47.01 & 41.97 \\
        SAN-SAW \cite{peng2022semantic} (CVPR'22)         & \secondmark{52.95} & \textbf{28.32} & 47.28 & \secondmark{42.85} \\
        DIRL \cite{xu2022dirl} (AAAI'22)                  & 51.80 & 26.50 & 46.52 & 41.61 \\
        SHADE \cite{zhao2022style} (ECCV'22)              & 50.95 & 27.62 & \secondmark{48.61} & 42.39 \\
        \rowcolor{gray!20} \textbf{Ours}     & \textbf{53.85} & 27.08 & \textbf{50.09} & \textbf{43.67} \\
        \bottomrule
    \end{tabular}}
    \caption{Mean IoU (\%, higher is better) comparison between DGSS methods trained on Cityscapes and evaluated on BDD100K (B), Synthia (S), and GTAV (G) with the ResNet-50 backbone.}
    \label{tab:cityscapes}
\vspace{-0.4cm}
\end{table}

\subsection{Domain Invariance Analysis}

\textbf{Quantitative Analysis.}
We propose a domain invariance quantitative analysis approach based on the Chamfer Distance.
Tab.\ref{tab:domain_sim} shows the results; the higher the metric, the better the model.
It shows that the proposed SRMA has superior domain invariance, especially at the regional and local levels. 
Meanwhile, SRMA remains competitive in global invariance compared to methods that adopt global randomization at the image (TLDR) or feature (SHADE) level.
More details about the proposed domain invariance quantitative analysis approach are described in Appendix \ref{sec:domain_invariance_details}.

\begin{table}[t!]
    \centering
    \scaletable{
    \begin{tabular}{l|ccc|ccc|ccc}
        \toprule
        \multirow{2}{*}{Methods}  & \multicolumn{3}{c|}{Global} & \multicolumn{3}{c|}{Local} & \multicolumn{3}{c}{Regional}  \\
            & C & B & M &   C & B & M &     C & B & M     \\
        \midrule
        SHADE \cite{zhao2022style} (ECCV'22)        & 0.909 & 0.894 & \secondmark{0.913} & \secondmark{0.582} & 0.566 & \secondmark{0.577} & 0.684 & 0.672 & 0.683 \\
        TLDR$^{\S}$ \cite{kim2023texture} (ICCV'23) & \textbf{0.914} & \textbf{0.913} & \textbf{0.915} & 0.574 & \secondmark{0.567} & 0.576 & \secondmark{0.736} & \secondmark{0.681} & \secondmark{0.729} \\
        \rowcolor{gray!20}
        \textbf{Ours}                               & \secondmark{0.911} & \secondmark{0.898} & \textbf{0.915} & \textbf{0.630} & \textbf{0.623} & \textbf{0.632} & \textbf{0.738} & \textbf{0.726} & \textbf{0.739}    \\
        \bottomrule
    \end{tabular}}
    \caption{
    Domain invariance quantitative analysis.
    Models are trained on GTAV using ResNet-50.
    }
    \label{tab:domain_sim}
    \vspace{-0.4cm}
\end{table}

\begin{figure}[t!]
    \centering
    \includegraphics[width=0.99\linewidth]{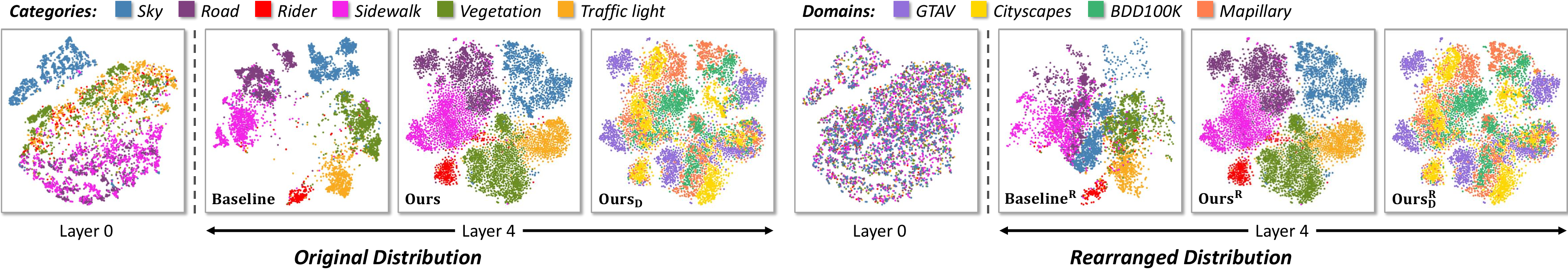}
    \vspace{-0.1cm}
    \caption{
    T-SNE visualization of features in different categories and domains extracted by the baseline model and our SRMA.
    Models are trained on GTAV using ResNet-50.
    }
    \label{fig:vis_layer4}
    \vspace{-0.4cm}
\end{figure}

\textbf{Qualitative Analysis.}
We utilize T-SNE to perform dimensionality reduction on the mean values of each semantic region to analyze the effect of SRM and the effectiveness of our proposed framework.
Fig.\ref{fig:vis_layer4} presents the original distribution (left) and semantic-regional-rearranged distribution (right) of shallow features (Layer 0).
It is obvious that SRM provides complex semantic regional attributes for the source domain, which significantly diversifies the source domain so as to facilitate the network to generalize to the variability of regional characteristics across domains.
The baseline model is capable of extracting distinguishable deep features by category from the original distribution (Baseline), but when faced with the rearranged distribution, the deep features become quite ambiguous (Baseline$^\text{R}$).
In contrast, our method extracts almost congruent deep representations (Ours \& Ours$^\text{R}$) from both shallow feature distributions, suggesting our method's robustness to varied semantic regional characteristics.
Moreover, we further represent the features of the various domains as points in different colors (Ours$_\text{D}$ \& Ours$_\text{D}^\text{R}$).
Our observation reveals that points of the different domains are well grouped by category, showing the capability of our method to obtain a domain-invariant representation and generalize to unseen target domains through training on a single source domain.

\subsection{Ablation Studies}

\begin{table}[t]
\begin{minipage}{.55\linewidth}
    \centering
    \scalebox{0.85}{
    \begin{tabular}{ccc|ccc|c}
        \toprule
        SRM     & $\mathcal{L}_{MLA}$ & $\mathcal{L}_{PC}$  & C     & B     & M     & Avg.  \\
        \midrule
         &    &     & 28.95 & 25.14 & 28.18 & 27.42 \\
        \Checkmark  &   &                                   & 42.39 & 35.78 & 39.89 & 39.35        \\
        \Checkmark  &  \Checkmark &                         & 44.13 & \secondmark{40.06} & \secondmark{45.08} & \secondmark{43.09}        \\
        \Checkmark  &   &  \Checkmark                       & \secondmark{45.36} & 39.01 & 41.56 & 41.98        \\
        \rowcolor{gray!20} \Checkmark  & \Checkmark  & \Checkmark  & \textbf{47.05} & \textbf{41.55} & \textbf{47.96} & \textbf{45.52}  \\
        \bottomrule
    \end{tabular}
    }
    \caption{
    Ablation studies for the main components of SRMA.
    Models are trained on GTAV using ResNet-50.
    }
    \label{tab:ablation}
\end{minipage}
\hspace{0.1cm}
\begin{minipage}{.42\linewidth}
    \centering
    \scalebox{0.85}{
    \begin{tabular}{l|ccc|c}
        \toprule
        Methods & C     & B     & M     & Avg.      \\
        \midrule
        w/o $\mathcal{A}_g$     & 45.81 & \secondmark{41.38} & 45.35 & 44.18 \\
        w/o $\mathcal{A}_r$     & \secondmark{46.48} & 41.11 & \secondmark{46.51} & \secondmark{44.70} \\
        w/o $\mathcal{A}_l$     & 45.13 & 39.84 & 42.20 & 42.39 \\
        \rowcolor{gray!20} \textbf{Ours}    & \textbf{47.05} & \textbf{41.55} & \textbf{47.96} & \textbf{45.52} \\
        \bottomrule
    \end{tabular}
    }
    \caption{
    Ablation studies for each level used in MLA.
    Models are trained on GTAV using ResNet-50.
    }
    \label{tab:level_mla}
\end{minipage}
\vspace{-0.4cm}
\end{table}

\noindent\textbf{Main components.}
As shown in Tab.\ref{tab:ablation}, we investigate how each main component contributes to our SRMA.
As the foundation of our framework, SRM improves the average mIoU by 11.93\% over the baseline without resorting to other losses.
This exhibits the ability of SRM to provide reliable and diverse randomized samples.
Moreover, $\mathcal{L}_{MLA}$ and $\mathcal{L}_{PC}$ provide different contributions to SRMA.
$\mathcal{L}_{MLA}$ provides consistent domain-invariant information, thus improving generalization performance.
$\mathcal{L}_{PC}$ further provides task-specific domain insensitivity, and in collaboration with $\mathcal{L}_{MLA}$, it augments the average mIoU by 6.17\% over using only SRM.

\begin{wrapfigure}{r}{7.0cm}
	\centering
    \vspace{-0.3cm}
        \begin{subfigure}[]{0.49\linewidth}
             \centering
             \includegraphics[width=\linewidth]{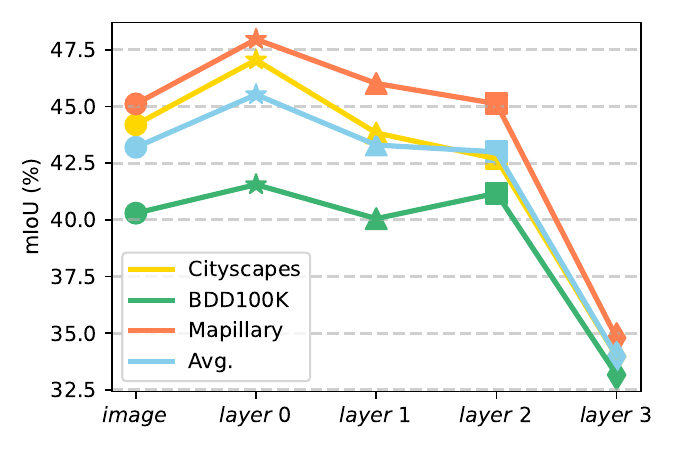}
             \caption{Location of SRM}  
             \label{fig:srm_layer}
         \end{subfigure} 
         \hfill
        \begin{subfigure}[]{0.49\linewidth}
             \centering
             \includegraphics[width=\linewidth]{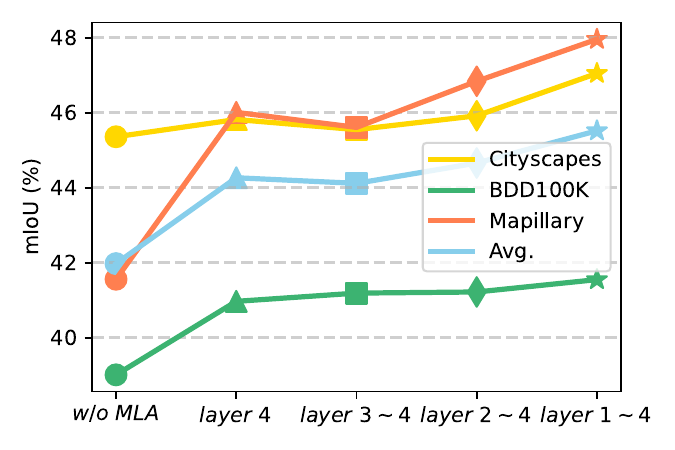} 
             \caption{Layers involved in MLA}  
             \label{fig:mla_layer}
         \end{subfigure}
         \hfill
         \vspace{-0.1cm}
	\caption{Change in performance with the location of SRM and the layers involved in MLA.}
    \vspace{-0.3cm}
	\label{fig:srm_mla_layer}
\end{wrapfigure}

\noindent\textbf{Analysis of SRM and MLA implementation locations.}
As shown in Fig.\ref{fig:srm_layer}, we first investigate the optimal location of SRM by applying it on images and the features extracted from layer $0\sim3$, respectively.
Obviously, performing SRM directly on images is inferior to processing shallow features.
Furthermore, as the layer deepens, the benefit from conducting SRM declines significantly.
This decline can be attributed to the style features contained in images are not clearly distinct, as well as the style information in deep features is gradually displaced by the content information. 
Therefore, we randomize the features extracted by layer $0$ to maximally diversify the semantic regional styles of the source domain.
Fig.\ref{fig:mla_layer} presents the ablation studies on the layers involved in MLA.
It turns out that involving all of the deep layers makes MLA more effective, reflecting the heuristic of constraining multiple layers.

\noindent\textbf{Effectiveness of each level in MLA.}
We exhibit the effects of each level contained in MLA in Tab.\ref{tab:level_mla}, which shows that the performance for unseen target domains decreases consistently when any level within MLA is excluded.
It demonstrates that each level in MLA plays an important role in helping the model understand domain-invariant attributes more comprehensively from different perspectives.
Additional ablation studies, network complexity, limitation and future work, segmentation qualitative comparison, visualization results, and analyses are provided in the Appendix.



\section{Conclusion}

This work is the first to focus on the phenomenon that different semantic regions have different characteristics across domains.
We proposed the Semantic-Rearrangement-based Multi-Level Alignment (SRMA) to mitigate the negative impact of this phenomenon on DGSS.
SRMA first included the Semantic Rearrangement Module (SRM), which was designed to diversify the characteristics of semantic regions in the source domain.
Subsequently, the Multi-Level Alignment (MLA) was proposed to learn domain-invariant representations from multiple levels with the help of such diversity, thus tackling the key challenge of learning the generalization capability of semantic segmentation.
SRMA has achieved state-of-the-art performance in multiple DGSS benchmarks, and extensive experiments have been conducted to demonstrate the superiority of SRMA.
Further visualizations and analyses have exhibited the robustness of SRMA to diverse semantic regional characteristics, indicating the remarkable capability of SRMA to generalize to the unseen target domains.

\bibliographystyle{ieee_fullname}
\bibliography{neurips_2024}

\newpage
\appendix




\renewcommand\thesection{A.\arabic{section}}
\renewcommand\thesubsection{A.\arabic{section}.\arabic{subsection}}
\renewcommand{\thetable}{A\arabic{table}}
\renewcommand{\thefigure}{A\arabic{figure}}

\setcounter{section}{0}
\setcounter{table}{0}
\setcounter{figure}{0}
\setcounter{equation}{0}

\newcommand{\rb}[1]{\rotatebox{90}{#1}}


\section{Qualitative Comparison}

\begin{figure}[h]
    \centering
    \includegraphics[width=0.99\linewidth]{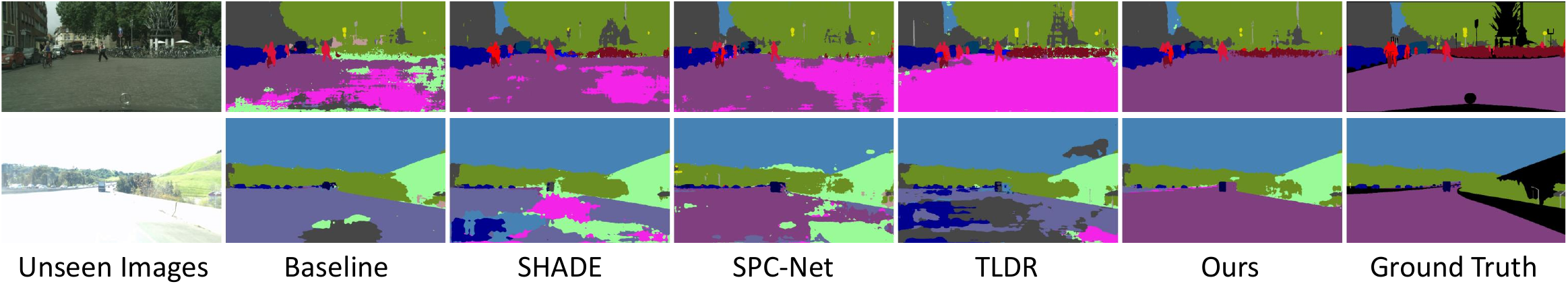}
    \vspace{-0.1cm}
    \caption{
    Qualitative results of existing DGSS methods and our SRMA on unseen domains.
    Models are trained on GTAV using ResNet-50.
    }
    \label{fig:qualitative_main}
\end{figure}

We compare the qualitative segmentation results between Baseline \cite{chen2018encoder}, SHADE \cite{zhao2022style}, SPC-Net \cite{huang2023style}, TLDR \cite{kim2023texture}, and our proposed SRMA.
The visualization of the results on unseen target domains is shown in Fig.\ref{fig:qualitative_main}.
It can be observed that our method yields prediction maps with fewer perturbations (line 1).
Our method also achieves reliable results for samples with extreme lighting conditions (line 2).
These results intuitively indicate the robustness of our method in real street segmentation scenarios with superior generalization performance.
Additional qualitative results are provided in Appendix \ref{sec:add_qual_res}.

\section{Experiments with Lightweight Backbones}
\label{sec:lightweight}

We further evaluate our SRMA using lightweight backbones (\textit{i.e.}, ShuffleNet \cite{ma2018shufflenet} and MobileNet \cite{sandler2018mobilenetv2}), the comparison results are shown in \ref{tab:lightweight}. 
In terms of both individual unseen target domains and average performance, our method shows consistently superior performances on lightweight backbones.
It makes a strong case for the superiority of our method while also demonstrating that our approach supports use in more practical scenarios.

\begin{table}[ht]
    \centering
    \scaletable{
    \begin{tabular}{l|ccc|c||ccc|c}
        \toprule
        \multirow{2}{*}{Methods}    & \multicolumn{4}{c||}{ShuffleNet} & \multicolumn{4}{c}{MobileNet} \\
            & C & B & M & Avg. & C & B & M & Avg. \\
        \midrule
            Baseline  & 25.56 & 22.17 & 28.60 & 25.44   & 30.14 & 27.66 & 27.07 & 28.29    \\
            IBN-Net \cite{pan2018two} (ECCV'18)               & 27.10 & 31.82 & 34.89 & 31.27   & 30.14 & 27.66 & 27.07 & 28.29 \\ 
            ISW \cite{choi2021robustnet} (CVPR'21)            & 30.98 & 32.06 & 35.31 & 32.78   & 30.86 & 30.05 & 30.67 & 30.53 \\
            DIRL \cite{xu2022dirl} (AAAI'22)                  & 31.88 & 32.57 & \secondmark{36.12} & 33.52  & 34.67 & 32.78 & \secondmark{34.31} & \secondmark{33.92} \\
            SHADE$^*$ \cite{zhao2022style} (ECCV'22)          & \secondmark{35.39} & \secondmark{32.98} & 35.10 & \secondmark{34.49}   & \secondmark{35.33} & \secondmark{32.89} & 31.18 & 33.13  \\
        \rowcolor{gray!20} \textbf{Ours}            & \textbf{36.59} & \textbf{33.56} & \textbf{36.41} & \textbf{35.52}     & \textbf{36.54} & \textbf{34.15} & \textbf{35.61} & \textbf{35.43} \\
        \bottomrule
    \end{tabular}}
    \caption{Mean IoU (\%, higher is better) comparison between existing DGSS methods trained on GTAV and evaluated on Cityscapes (C), BDD100K (B), and Mapillary (M) with lightweight backbones. 
    The best and second best results are \textbf{highlighted} and \secondmark{underlined}, respectively. 
    $^{\S}$ denotes using extra data during training.
    The results of previous methods come from their public reports.
    $^*$ indicates that we re-implement them with a replaced backbone on their publicly available implementations, while the other results come from their public reports.
    }
    \label{tab:lightweight}
    \vspace{-0.4cm}
\end{table}

\section{Experiments with Different Architectures}
\label{sec:m2f}

We introduce an advanced architecture Mask2Former \cite{cheng2022masked}, in place of DeepLabV3+ \cite{chen2017deeplab} used in the main experiments to validate the robustness of SRMA to semantic segmentation frameworks.
We conduct our experiments by replacing the decoder and task-specific loss $\mathcal{L}_{task}$ with those used in Mask2Former.
Mask2Former introduces 0.5M additional parameters while leading to competitive performance, as shown in Tab.\ref{tab:m2f}.
It suggests SRMA's effectiveness in advanced architecture, demonstrating that our proposed method is highly generalizable for semantic segmentation tasks.

\vspace{-0.3cm} \begin{table}[ht]
    \centering
    \scaletable{
    \begin{tabular}{l|cccc}
        \toprule
        Methods & ~~~~C~~~~ & ~~~~B~~~~ & ~~~~M~~~~ & ~~~~Avg.~~~~  \\
        \midrule
        Ours (DeepLabV3+)      & 47.05 & 41.55 & \textbf{47.96} & 45.52         \\
        Ours (Mask2Former)     & \textbf{47.62} & \textbf{43.76} & 46.62 & \textbf{46.00}         \\
        \bottomrule
    \end{tabular}}
    \caption{Experiments using different architectures. Models are trained on GTAV.}
    \label{tab:m2f}
    \vspace{-0.4cm}
\end{table}

\section{Details of Domain Invariance Quantitative Analysis}
\label{sec:domain_invariance_details}

Inspired by the domain similarity metrics in previous works \cite{cui2018large}, we propose that it is feasible to measure the domain invariance of a model using the feature distances across domains in a normalized feature space.
Specifically, for the source domain $\mathcal{S}$ and the target domain $\mathcal{T}$, we extract sample features for both domains using the same model and compare the distance between these two feature sets.
If the distance between the two domains in the feature space is small, it suggests that the model extracts similar features for the samples from these two domains, which in turn responds that the model has strong domain invariance between $\mathcal{S}$ and $\mathcal{T}$.

To accomplish this goal, we use the Chamfer Distance as the distance metric for features across domains and a monotonically decreasing exponential function as the similarity function.
Chamfer Distance is typically utilized to measure the spatial distance between two point sets and is often used in 3D reconstruction evaluations \cite{wang2021neus,zhang2024ddf,zhang2023moho}.
It can be obtained as follows:
\begin{equation}
    d_{\mathcal{CD}}(\mathcal{S}, \mathcal{T}) = \frac{
    \sum_{s \in \mathcal{S}} \min_{t \in \mathcal{T}} \Vert s - t \Vert
    }{2 \vert \mathcal{S} \vert } + \frac{ 
    \sum_{t \in \mathcal{T}} \min_{s \in \mathcal{S}} \Vert t - s \Vert
    }{2 \vert \mathcal{T} \vert }.
\end{equation}
After obtaining the domain distance, we calculate the domain invariance between these two domains as follows:
\begin{equation}
    \mathcal{I}(\mathcal{S}, \mathcal{T}) = e^{-\gamma d_{\mathcal{CD}}(\mathcal{S}, \mathcal{T})},
\end{equation}
where $\gamma$ is set to 0.01 in our experiments.

Moreover, we choose layer 4 features which have more content information for calculating the domain distance.
Since the features extracted from different models exist in different feature spaces, to quantitatively compare the domain invariance of different models, we furthermore standardize the features of layer 4 to eliminate the scale difference between different feature spaces.
Specifically, for each model, we obtain the mean and standard deviation of the sampled local features and use them to normalize features at each level.

We compare domain invariance at the global, local, and regional levels, respectively.
For the global level, we take the global mean of features as the global feature for that sample.
For the local level, each pixel feature is an individual local feature.
For the region level, we adopt semantic average pooling to obtain the respective mean value of each semantic region on the sample and use them as region features respectively.
We obtain domain invariance at the local and regional levels by first calculating the respective domain invariance for each semantic category separately and then fetching their mean values.
In turn, we do not consider semantic categories when calculating the global domain invariance.

We take the average of 10 trials as the result.
In each trial, we respectively sample 300 features at the global, local, and regional levels for each model and domain.
The algorithm for a single trial calculating the domain invariance between the source domain $\mathcal{S}$ and the target domain $\mathcal{T}$ is shown in Algorithm \ref{alg:domain_inv}.

\begin{algorithm}[ht]
        \SetAlgoLined
	\SetKwInOut{Input}{Input}
	\SetKwInOut{Output}{Output} 
	\SetKwInOut{Parameter}{Param}
        \SetKwFunction{Sample}{sample}

    \Input{Source domain feature set $\mathcal{S}$. \\
            Target domain feature set $\mathcal{T}$. \\
    }
    \Parameter{Exponential coefficient $\gamma$.\\}
    \Output{Domain invariance \{$\mathcal{I}_g$, $\mathcal{I}_l$, $\mathcal{I}_r$\}.}

    $\mu_{\mathcal{S}} \gets \mu(\mathcal{S})$ \\
    $\sigma_{\mathcal{S}} \gets \sigma(\mathcal{S})$ \\
    
    $\mathcal{S}_g \gets \{\frac{GAP(s) - \mu_{\mathcal{S}}}{\sigma_{\mathcal{S}}} ~\vert~ s \in \mathcal{S}\}$ \\
    $\mathcal{T}_g \gets \{\frac{GAP(t) - \mu_{\mathcal{S}}}{\sigma_{\mathcal{S}}} ~\vert~ t \in \mathcal{T}\}$ \\

    $\mathcal{S}_l \gets \{ \frac{s_{h,w} - \mu_{\mathcal{S}}}{\sigma_{\mathcal{S}}} ~\vert~ s_{h,w} \in \mathcal{S} \} $ \\
    $\mathcal{T}_l \gets \{\frac{t_{h,w} - \mu_{\mathcal{S}}}{\sigma_{\mathcal{S}}} ~\vert~ t_{h,w} \in \mathcal{T}\} $ \\
    
    $\mathcal{S}_r \gets \{\frac{SAP(s, c) - \mu_{\mathcal{S}}}{\sigma_{\mathcal{S}}} ~\vert~ s \in \mathcal{S}, c \in \{1, ..., C\}\} $ \\
    $\mathcal{T}_r \gets \{\frac{SAP(t, c) - \mu_{\mathcal{S}}}{\sigma_{\mathcal{S}}} ~\vert~ t \in \mathcal{T}, c \in \{1, ..., C\}\} $ \\
    
    \ForEach{$j \in \{g, l, r\}$}{
        $\mathcal{S}_j \gets \Sample(\mathcal{S}_j)$ \\
        $\mathcal{T}_j \gets \Sample(\mathcal{T}_j)$ \\
        $d_j \gets \frac{
            \sum_{s \in \mathcal{S}_j} \min_{t \in \mathcal{T}_j} \Vert s - t \Vert
            }{2 \vert \mathcal{S}_j \vert } + \frac{ 
            \sum_{t \in \mathcal{T}_j} \min_{s \in \mathcal{S}_j} \Vert t - s \Vert
            }{2 \vert \mathcal{T}_j \vert }$ \\
        $\mathcal{I}_j \gets e^{-\gamma d_j}$ \\
    } 
    
    \caption{Domain invariance quantitative analysis in a single trial.}
    \label{alg:domain_inv}
\end{algorithm}

\section{Additional Ablation Studies}
\label{sec:add_abla}

\subsection{Ablation Studies on Fixing Layer 0}
During the training stage, our framework fixes layer 0 of the backbone with ImageNet \cite{deng2009imagenet} pre-trained parameters in order to avoid overfitting the source domain.
Since the purpose of SRM is to generate new samples with randomized regional characteristics in the shallow feature space after layer 0, the samples generated by SRM do not participate in the gradient flow of the original samples.
If layer 0 is trainable, it can only learn representations from the original samples, which can easily lead to overfitting the source domain.

As shown in Tab.\ref{tab:fix0_abla}, training the baseline model with a fixed layer 0 (Baseline w F.L.0) does not significantly affect its performance, suggesting that reducing the number of trainable parameters by such a small amount does not affect the model's learning of task-specific knowledge.
Moreover, the performance of our method degrades when layer 0 is not fixed (Ours w/o F.L.0).
It suggests that a layer 0 that overfits the source domain negatively impacts the effectiveness of SRM, which in turn prevents the model from obtaining domain-invariant representations when tested on unseen target domains.
Therefore, we fixed the parameters of layer 0 in the training stage to avoid this negative phenomenon.

\begin{table}[ht]
    \centering
    \scaletable{
    \begin{tabular}{l|ccc|c}
        \toprule
        Methods~~~~~~~~~~~~~~ & ~~~~C~~~~  & ~~~~B~~~~   & ~~~~M~~~~   & ~~~~Avg.~~~~    \\
        \midrule
        Baseline            & 28.95 & 25.14 & 28.18 & 27.42     \\
        Baseline w F.L.0    & 29.04 & 25.74 & 28.25 & 27.68     \\
        \midrule
        Ours w/o F.L.0      & 45.03 & 41.07 & 44.40 & 43.50     \\
        \rowcolor{gray!20} \textbf{Ours} & \textbf{47.05} & \textbf{41.55} & \textbf{47.96} & \textbf{45.52}        \\
        \bottomrule
    \end{tabular}}
    \caption{
    Analysis of fixing layer 0 in the feature extractor.
    F.L.0 indicates the fixing layer 0.
    Models are trained on GTAV and evaluated on Cityscapes, BDD100K, and Mapillary with ResNet-50.
    }
    \label{tab:fix0_abla}
    \vspace{-0.4cm}
\end{table}

\subsection{Ablation Studies on the Style Elimination}

We adopt the style elimination in our framework that attempts to eliminate global styles for shallow features.
With the help of style elimination, the following deep layers in the network will only fetch features with a global mean of 0 and a global standard deviation of 1.

Tab.\ref{tab:se_abla} shows the ablation studies on the style elimination.
In the absence of SRM and additional losses, the model with a style elimination (Baseline w S.E.) achieves a significant improvement over the baseline, which roughly reflects the contribution of the style elimination to DGSS. 
Meanwhile, it suggests that style elimination is capable of learning domain-invariant representations at the global level by simply eliminating the global style of shallow features.
Furthermore, taking away the style elimination in our framework (Ours w/o S.E.) brings down the overall performance, demonstrating its positive role in our framework.

This light modification of network architecture arises after analyzing previous works applying global style randomization to shallow features \cite{lee2022wildnet,zhao2022style}.
These approaches store additional global statistics to provide a reference for the global style transfer in the training stage.
Such randomization methods result in samples with different global appearances, but it seems that such global variance is of little significance for the generalization capability as it can be filtered out via IN so as not to have an impact on generalized segmentation \cite{pan2018two}.
Since randomization of the global style is difficult to be exhaustive while shallow features are not content-sensitive, we hypothesize that ignoring the global style of shallow features can better mitigate the source-target domain gap while avoiding semantic information attenuation.
Therefore, instead of further randomization, we perform style elimination at the global level of shallow features extracted by layer $0$, which filters out the complex global appearances while preserving the content to help subsequent learning of generalized segmentation.
As a result, it facilitates MLA to learn domain-invariant representations.

\begin{table}[ht]
    \centering
    \scaletable{
    \begin{tabular}{l|ccc|c}
        \toprule
        Methods~~~~~~~~~~~~~~ & ~~~~C~~~~  & ~~~~B~~~~   & ~~~~M~~~~   & ~~~~Avg.~~~~    \\
        \midrule
        Baseline            & 28.95 & 25.14 & 28.18 & 27.42     \\
        Baseline w S.E.~~   & 38.01 & 33.82 & 37.63 & 36.49     \\
        \midrule
        Ours w/o S.E.       & 46.45 & 40.32 & 45.49 & 44.09     \\
        \rowcolor{gray!20} \textbf{Ours} & \textbf{47.05} & \textbf{41.55} & \textbf{47.96} & \textbf{45.52}        \\
        \bottomrule
    \end{tabular}}
    \caption{
    Ablation studies for the style elimination.
    S.E. indicates the style elimination.
    Models are trained on GTAV and evaluated on Cityscapes, BDD100K, and Mapillary with ResNet-50.
    }
    \label{tab:se_abla}
    \vspace{-0.4cm}
\end{table}

\subsection{Hyper-parameter Analysis}

Tab.\ref{tab:hp_ablation} shows the ablation studies for hyper-parameters of our SRMA.
We choose the hyper-parameters empirically depending on the properties of modules or the optimization process.
To help models gradually understand the domain-neutral knowledge, we intuitively choose weight combinations that increase with the depth of layers as $\lambda_{MLA}$, based on the decrease in style information and increase in content information after the layer deepens as \cite{pan2018two} suggests.
We set $\lambda_{PC}$ to 10 based on the scale of $\mathcal{L}_{PC}$ to balance it with the scale of $\mathcal{L}_{MLA}$ for optimization.
$Dir$ refers to the parameter of the Dirichlet distribution used for sampling random weights for SRM. 
The closer $Dir$ to 1, the more uniform the sampled weights are, resulting in the synthetic distribution (\textit{Sy}) to average over all semantic distributions (\textit{Se}); 
the closer $Dir$ to 0, the more heavy-tailed the sampled weights are, resulting in the \textit{Sy} to approximate a particular \textit{Se}. 
Our choice of $2^{-6}$ is for producing more diverse \textit{Sy} by linearly combining real \textit{Se}.

\begin{table}[ht]
    \centering
    \scaletable{
    \begin{tabular}{l|c|ccc|c}
        \toprule
        Hyper-parameters     & Value & ~~~~C~~~~ & ~~~~B~~~~  & ~~~~M~~~~  & ~~~~Avg.~~~~  \\
        \midrule
        \multirow{2}{*}{$\lambda_{MLA}$}    & [1, 0.8, 0.6, 0.4]    & 46.53 & 40.98 & 47.52 & 45.01 \\
                                            & [1, ~~~1, ~~~1, ~~~1] & \secondmark{46.96} & 41.23 & 47.37 & 45.19 \\
        \midrule
        \multirow{2}{*}{$\lambda_{PC}$}     & 5         & 47.02 & 40.32 & 47.48 & 44.94 \\
                                            & 15        & 46.88 & 41.43 & \secondmark{47.76} & \secondmark{45.36} \\
        \midrule
        \multirow{2}{*}{$Dir$}              & $2^{-4}$  & 46.09 & 41.08 & 46.76 & 44.64 \\
                                            & 0         & 46.79 & 40.98 & 47.51 & 45.09 \\
        \midrule
        \rowcolor{gray!20} 
        \textbf{Ours}    & Eq. 11 & \textbf{47.05} & \textbf{41.55} & \textbf{47.96} & \textbf{45.52} \\
        \bottomrule
    \end{tabular}}
    \caption{
    Ablation studies for hyper-parameter selection.
    $\lambda_{MLA}$ and $\lambda_{PC}$ denote the weights of the overall training objective.
    $Dir$ indicates the parameter of the Dirichlet distribution utilized in SRM. 
    Models are trained on GTAV and evaluated on Cityscapes, BDD100K, and Mapillary with ResNet-50.
    }
    \label{tab:hp_ablation}
    \vspace{-0.4cm}
\end{table}

\section{Complexity of Networks}
\label{sec:complexity}
We further compare the complexity of our method with existing DGSS methods in Tab.\ref{tab:complexity}.
The number of parameters, the number of floating-point operation per second (FLOPs), and the inference time are used as comparison metrics for the computational cost.
Compared to the baseline, our method requires no extra parameters and is competitive in terms of computational complexity as well as inference time.
Some of the previous approaches introduce additional parameters (\textit{e.g.}, PintheMem \cite{kim2022pin}, DIRL \cite{xu2022dirl}, SPC-Net \cite{huang2023style}), include significant extra operations (\textit{e.g.}, ISW \cite{choi2021robustnet}, SAN-SAW \cite{peng2022semantic}), or optimize external auxiliary models (\textit{e.g.}, TLDR \cite{kim2023texture}).
In contrast, our model does not require additional parameters, while no auxiliary data or model is needed for training.
This comparison demonstrates that our method is highly cost-effective while exhibiting excellent performance.

\begin{table}[ht]
    \centering
    \scaletable{
    \begin{tabular}{l|ccc}
        \toprule
        ~~Methods~~ & \# of Params $\downarrow$  & FLOPs (G) $\downarrow$    & Times (ms) $\downarrow$    \\
        \midrule
        Baseline    & 45.08M        & 277.45 & 67.47       \\
        IBN-Net \cite{pan2018two} (ECCV'18)             & 45.08M        & 277.53 & 68.94         \\
        ISW \cite{choi2021robustnet} (CVPR'21)          & 45.08M        & 277.49 & 69.46         \\
        PintheMem \cite{kim2022pin} (CVPR'22)           & 45.28M        & 278.03 & 68.24         \\
        SAN-SAW \cite{peng2022semantic} (CVPR'22)       & 25.63M        & 422.48 & 180.10        \\
        DIRL \cite{xu2022dirl} (AAAI'22)                & 45.41M        & -      & -       \\
        SHADE \cite{zhao2022style} (ECCV'22)            & 45.08M        & 277.45 & 67.72        \\
        SPC-Net \cite{huang2023style} (CVPR'23)         & 45.22M        & -      & 94.59         \\
        TLDR$^\S$ \cite{kim2023texture} (ICCV'23)       & -             & -      & 153.82      \\
        \rowcolor{gray!20} \textbf{Ours}                & 45.08M        & 277.45 & 67.79         \\
        \bottomrule
    \end{tabular}}
    \caption{Computational cost comparison of ResNet-50 based model. 
    The times are tested with the image size of 2048$\times$1024 on one NVIDIA RTX 3090 GPU. 
    We averaged the inference time over 500 trials.
    FLOPs are carried out using \textit{ptflops}.}
    \label{tab:complexity}
    \vspace{-0.4cm}
\end{table}

\section{Per-category mIoU}

Tab.\ref{tab:per-category} provides the per-category mean intersection over union (mIoU).
We list the per-category performance of our proposed SRMA on Cityscapes \cite{cordts2016cityscapes}, BDD100K \cite{yu2020bdd100k}, and Mapillary \cite{neuhold2017mapillary} datasets, where the model is trained on GTAV \cite{richter2016playing} with the ResNet-50 \cite{he2016deep} backbone.

\begin{table*}[ht]
    \centering
    \scalebox{0.6}{
    \begin{tabular}{l|ccccccccccccccccccc|c}
        \toprule
        Datasets    & \rb{Road} & \rb{S.walk} & \rb{Build.} & \rb{Wall} & \rb{Fence} & \rb{Pole} & \rb{Tr. Light} & \rb{Tr. Sign} & \rb{Veget.} & \rb{Terrain} & \rb{Sky} & \rb{Person} & \rb{Rider} & \rb{Car} & \rb{Truck} & \rb{Bus} & \rb{Train} & \rb{Motor.} & \rb{Bicycle} & Avg.      \\
        \midrule
        Cityscapes  & 88.0 & 43.7 & 82.7 & 37.8 & 34.2 & 36.9 & 36.5 & 30.9 & 85.2 & 30.6 & 73.2 & 65.7 & 29.3 & 87.8 & 29.6 & 34.0 & 12.2 & 26.2 & 29.5 & 47.05    \\
        BDD100K     & 84.3 & 40.4 & 74.5 & 13.3 & 30.3 & 38.4 & 37.7 & 23.8 & 72.2 & 24.5 & 84.3 & 53.7 & 13.9 & 80.1 & 26.7 & 38.0 & - & 37.6 & 15.9 & 41.55    \\
        Mapillary   & 72.4 & 43.8 & 73.7 & 26.9 & 31.2 & 44.9 & 46.0 & 54.6 & 74.4 & 29.8 & 76.4 & 64.6 & 31.6 & 84.5 & 43.4 & 33.1 & 12.4 & 33.7 & 34.0 & 47.96    \\
        \bottomrule
    \end{tabular}}
    \caption{
    Per-category mIoU (\%) on three real-world datasets. 
    The model is trained on GTAV and evaluated on Cityscapes, BDD100K, and Mapillary with the ResNet-50 backbone.
    }
    \label{tab:per-category}
    \vspace{-0.4cm}
\end{table*}

\section{Analysis of the Domain-neutral Knowledge}

Our proposed MLA employs an ImageNet \cite{deng2009imagenet} pre-trained feature extractor to provide domain-neutral knowledge for alignments.
The introduction of domain-neutral knowledge is inspired by previous works \cite{hoyer2022daformer,zhao2022style}.
These methods apply the ImageNet pre-trained model to constrain the feature consistency of patch-level features in the \textit{thing} categories, thus allowing the models to obtain features closer to the real world in these categories.
Inspired by these methods, we introduce the domain-neutral knowledge contained in the ImageNet pre-trained model.
Unlike them, MLA does not only use domain-neutral knowledge to learn local expressions but also aligns at the global and regional levels with its help, thus helping the model to learn more comprehensive domain-invariant attributes.
Moreover, MLA does not only constrain the category \textit{thing}.
MLA leverages the domain-neutral knowledge without concerning categories to help the model learn more complete domain-invariant semantic information.

Furthermore, our departure point for using the pre-trained model is different from previous works as well.
The purpose of previous work introducing the ImageNet pre-trained model is to make some particular categories of features extracted by the segmentation model more similar to real-world features, thus improving the domain invariance for migration to the real-world domain on these categories.
On the contrary, our goal is to constrain the model to behave domain-invariant in its overall representation.
Therefore, we constrain all categories with the help of domain-neutral knowledge.
In addition, the model is allowed to gradually understand the domain-neutral knowledge by applying MLA with progressively increasing weights to multiple deep layers, so as to extract representations that have domain-invariant attributes at multiple levels.
In this work, we use the ImageNet pre-trained model that is accessible in all segmentation tasks in order to avoid the pipeline being too complex and thus unnecessarily consuming computational costs, as well as for fair comparisons.

\section{Limitation and Future Work}
The limitation of SRMA is its reliance on fine-grained semantic segmentation annotations for semantic rearrangement and regional level alignment.
In the absence of pixel-level annotations, the present framework cannot distinguish semantic information in different regions.

With the development of visual-language models and open-world segmentation techniques, some approaches such as Segment Anything\cite{Kirillov_2023_ICCV} can provide an approximate segmentation map for image processing tasks.
With the help of these methods, we can further migrate our framework to generalization applications for tasks such as image classification, object detection, instance segmentation, and so on.
We leave it as an interesting and meaningful future work.

\section{Visualization of Semantic-regional-rearranged Samples}

\begin{figure*}
    \centering
    \includegraphics[width=0.99\linewidth]{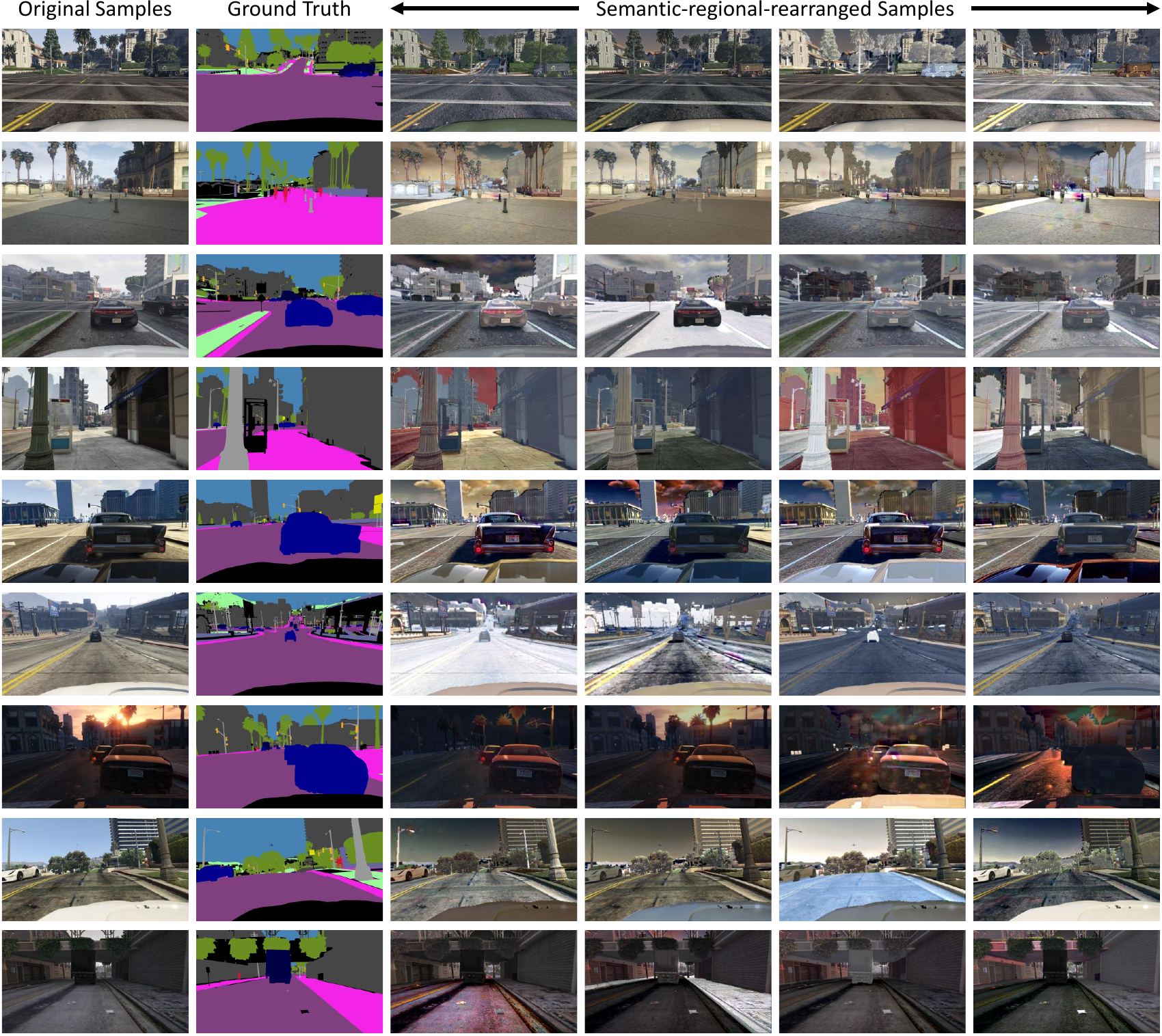}
    \vspace{-0.1cm}
    \caption{
    Visualization of semantic-regional-rearranged samples.
    These samples are obtained from GTAV.
    The randomized images shown above are reconstructed from the rearranged shallow features via a decoder used only for visualization.
    }
    \label{fig:srm_vis}
\end{figure*}

In Fig.\ref{fig:srm_vis}, we visualize the semantic-regional-rearranged samples with a reconstruction decoder.
Since SRM randomizes the semantic regional styles in the shallow feature space, intuitively the reconstructed samples are obviously different from the original samples in each semantic region.
Visually, the randomized samples generated by SRM contain diverse semantic region characteristics.
For example, regions in the \textit{road} and  \textit{sidewalk} categories have significantly different visual characteristics in different randomized samples that come from the same original sample.
SRM offers a broad range of brightness (brighter or darker) and color (bluer, greener, ...) for both categories.
Moreover, it can be observed that the \textit{sky}, \textit{vegetation}, and other categories also acquire extensive visual characteristics with the help of SRM.
It demonstrates that SRM is capable of fully diversifying the semantic regional characteristics of the source domain.

\section{Additional Qualitative Results}
\label{sec:add_qual_res}
\begin{figure*}
    \centering
    \includegraphics[width=0.99\linewidth]{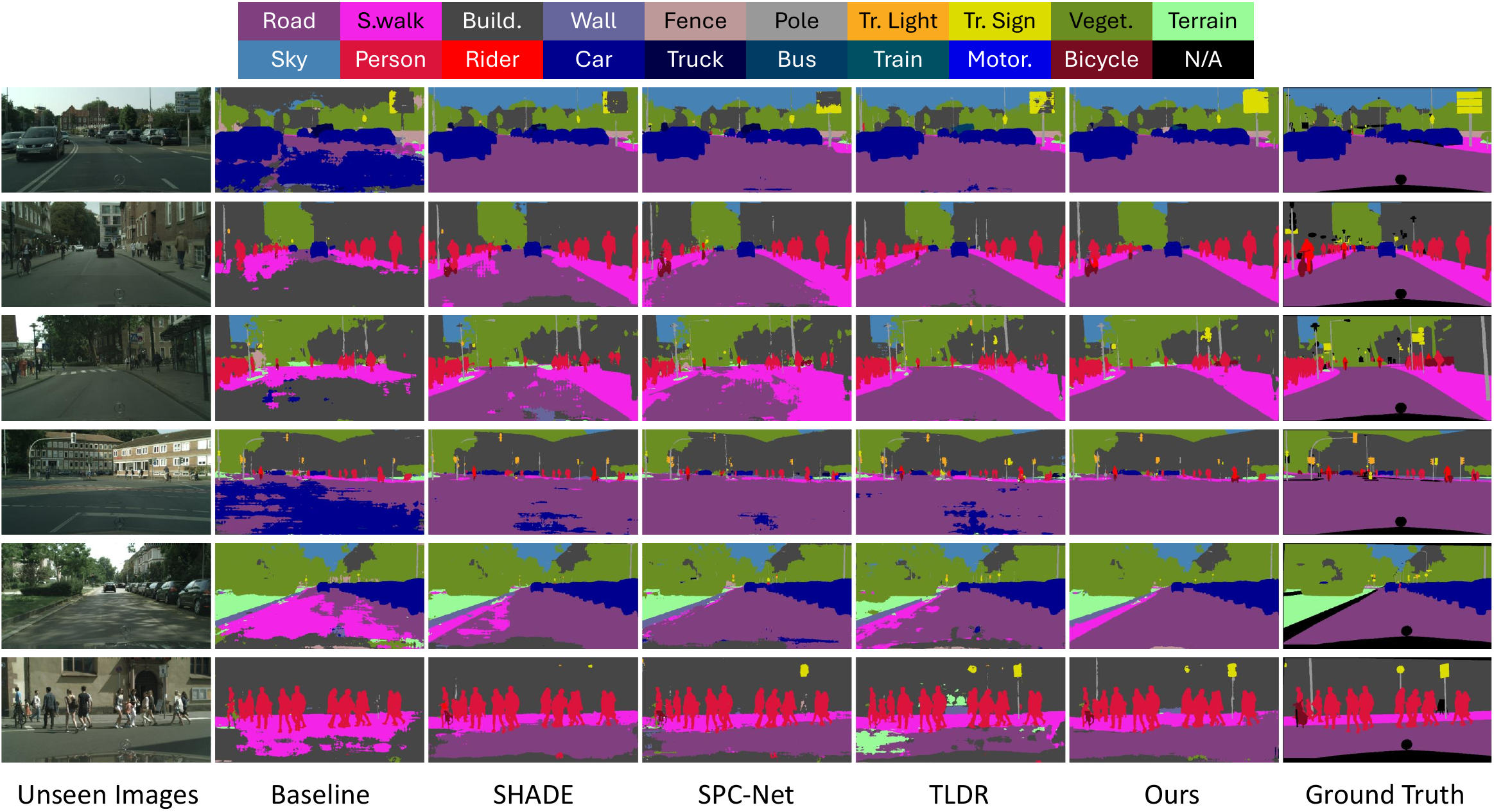}
    \vspace{-0.1cm}
    \caption{
    Qualitative results of DGSS methods and our SRMA on Cityscapes.
    Models are trained on GTAV using ResNet-50.
    }
    \label{fig:qualitative_city}
    \vspace{1.5cm}
    \centering
    \includegraphics[width=0.99\linewidth]{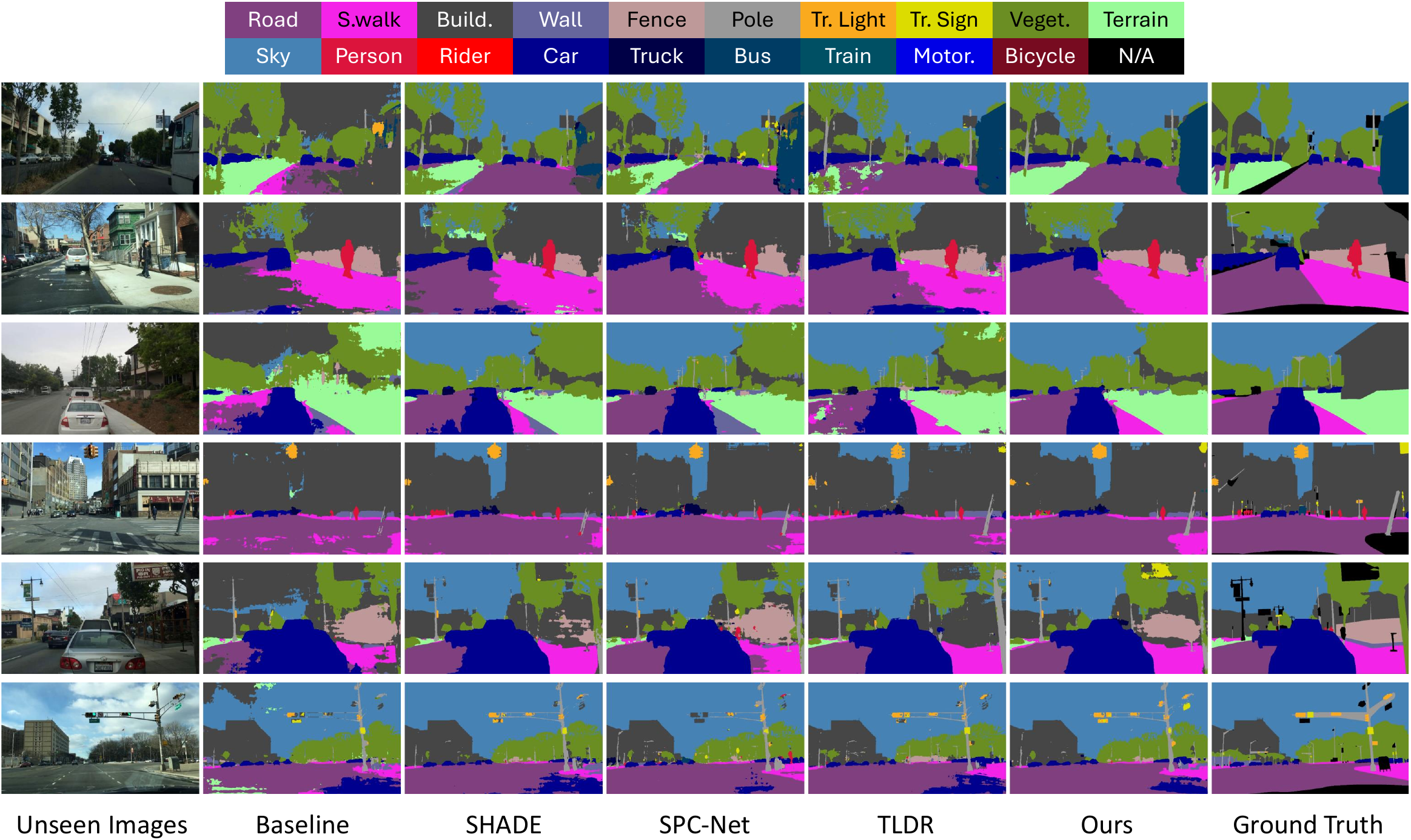}
    \vspace{-0.1cm}
    \caption{
    Qualitative results of DGSS methods and our SRMA on BDD100K.
    Models are trained on GTAV using ResNet-50.
    }
    \label{fig:qualitative_bdd}
\end{figure*}

\begin{figure*}
    \centering
    \includegraphics[width=0.99\linewidth]{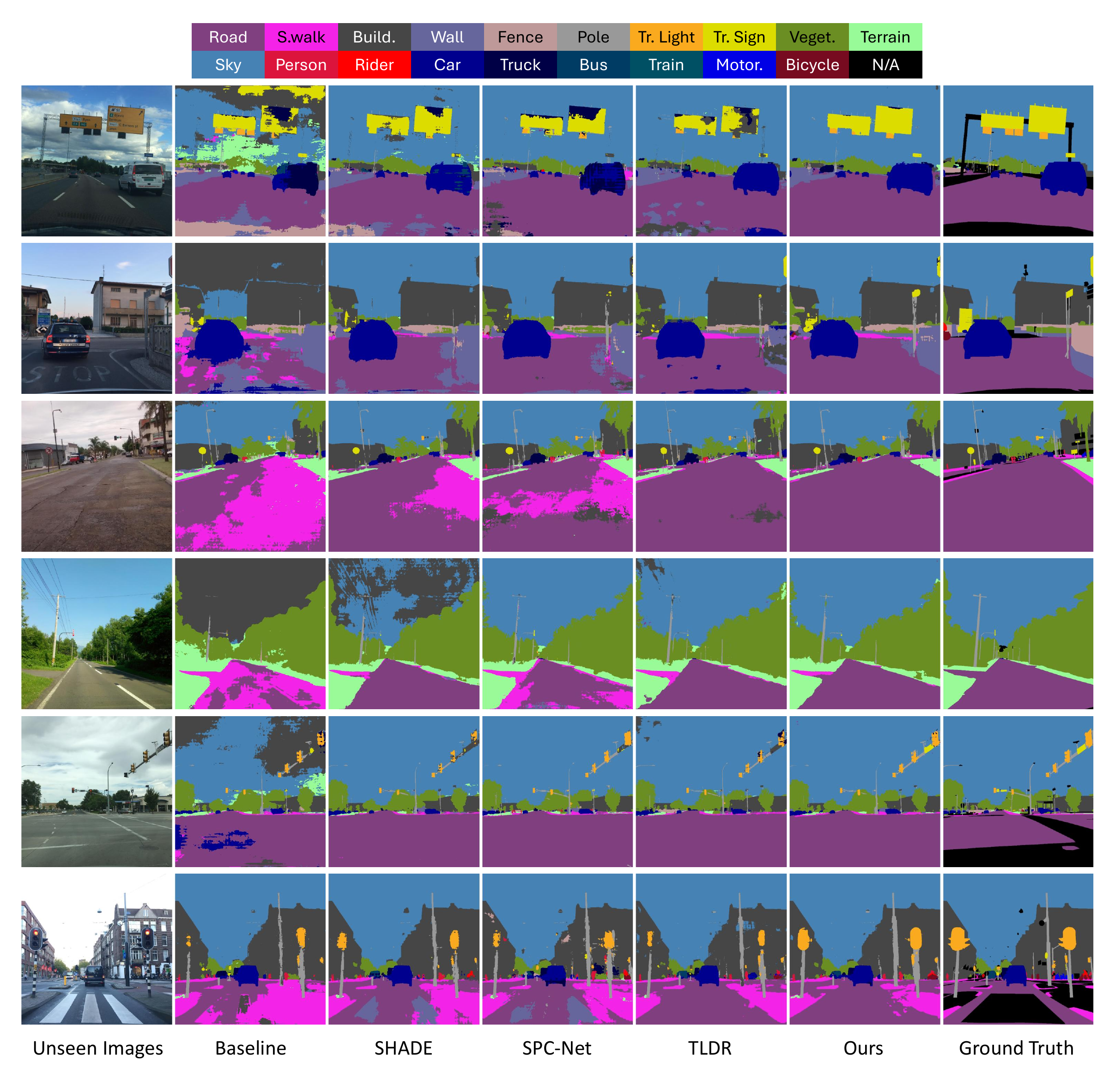}
    \vspace{-0.1cm}
    \caption{
    Qualitative results of DGSS methods and our SRMA on Mapillary.
    Models are trained on GTAV using ResNet-50.
    }
    \label{fig:qualitative_map}
\end{figure*}

In this section, we additionally compare the qualitative results on Cityscapes \cite{cordts2016cityscapes}, BDD100K \cite{yu2020bdd100k}, and Mapillary \cite{neuhold2017mapillary}, as shown in Fig.\ref{fig:qualitative_city}, Fig.\ref{fig:qualitative_bdd}, and Fig.\ref{fig:qualitative_map}, respectively.
Our method maintains robust predictions in the face of categories with large cross-domain gaps (\textit{e.g.}, \textit{traffic sign}, \textit{terrain}, and \textit{sidewalk}), demonstrating the superior capability of our approach to tackling the source-target domain gap and the robustness of the various regional characteristics across domains for individual semantic regions.

\end{document}